\documentclass[final]{cvpr}

\usepackage{times}
\usepackage{epsfig}
\usepackage{graphicx}
\usepackage{amsmath}
\usepackage{amssymb}

\usepackage{fancyhdr}

\usepackage{arydshln}
\usepackage{xcolor}
\definecolor{kitgreen}{RGB}{1,191,166}
\definecolor{lightblue}{RGB}{185,198,229}
\definecolor{red}{RGB}{230,10,30}
\definecolor{green}{RGB}{0, 128, 0}
\definecolor{blue}{RGB}{0, 0, 255}
\definecolor{gray}{RGB}{150, 150, 150}

\usepackage[pagebackref=true,breaklinks=true,colorlinks,bookmarks=false]{hyperref}

\usepackage{colortbl}
\usepackage{multirow}
\definecolor{lightsteelblue}{RGB}{176,196,222}
\definecolor{lightsteelred}{RGB}{230,176,160}
\definecolor{lightsteellila}{RGB}{175,181,224}
\definecolor{lightsteelgreen}{RGB}{182,214,207}

\def\cvprPaperID{4326} 
\def\confYear{CVPR 2021}

\begin{document}

\title{Every Annotation Counts: Multi-label Deep Supervision\\ for Medical Image Segmentation}


\author{Simon Rei{\ss}$^{1,2}$
~Constantin Seibold$^1$
~Alexander Freytag$^{2}$
~Erik Rodner$^{2,3}$
~Rainer Stiefelhagen$^1$\\
\normalsize
$^1$Karlsruhe Institute of Technology
\normalsize
~$^2$Carl Zeiss AG
\normalsize
~$^3$University of Applied Sciences Berlin\\
{\tt\small \{simon.reiss,constantin.seibold,rainer.stiefelhagen\}@kit.edu}
}

\maketitle
\thispagestyle{fancy}
\fancyhead[CE,CO]{\large In Proceedings of the IEEE Conference on Computer Vision and Pattern Recognition (CVPR), 2021}


\begin{abstract}
Pixel-wise segmentation is one of the most data and annotation hungry tasks in our field.
Providing representative and accurate annotations is often mission-critical especially for challenging medical applications.
In this paper, we propose a semi-weakly supervised segmentation algorithm to overcome this barrier. 
Our approach is based on a new formulation of deep supervision and student-teacher model and allows for easy integration of different supervision signals. 
In contrast to previous work, we show that care has to be taken how deep supervision is integrated in lower layers and we present \emph{multi-label deep supervision} as the most important secret ingredient for success.
With our novel training regime for segmentation that flexibly makes use of images that are either fully labeled, marked with bounding boxes, just global labels, or not at all, we are able to cut the requirement for expensive labels by $94.22\%$ -- narrowing the gap to the best fully supervised baseline to only $5\%$ mean IoU.
Our approach is validated by extensive experiments on retinal fluid segmentation and we provide an in-depth analysis of the anticipated effect each annotation type can have in boosting segmentation performance.

\end{abstract}

\section{Introduction}

\begin{figure}[t]
  \includegraphics[width=\linewidth]{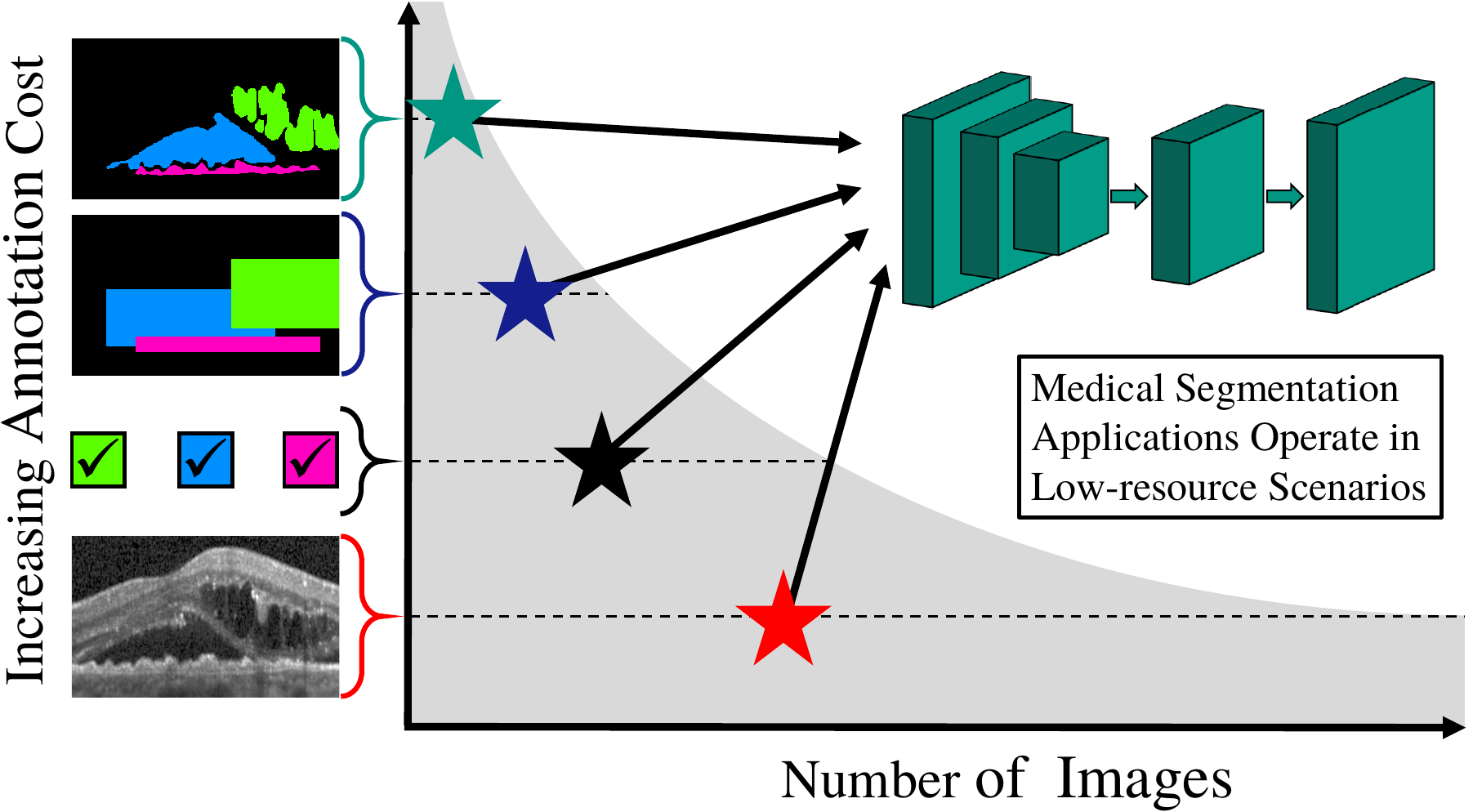}
  \caption{Annotations for segmentation are costly, especially when experts need to provide them. We show how our semi-weakly semantic segmentation method 
  can use different annotation types and how the recognition performance benefits from them.}
  \label{fig:overview}
\end{figure}
\noindent Medical imaging tools have become a central part in modern health care. In direct consequence, hundreds of millions of 2D- and 3D-images are recorded per year~\cite{Statista2019}. With this ever-growing number, medical personnel has become increasingly entangled in their evaluation. Deep learning with the use of artificial neural networks has found use to alleviate the necessary effort needed for interpretation of such images. However, training  these models requires enormous amounts of annotated data, especially in case of semantic segmentation. The annotation process for this task is already extremely difficult in real-world scenarios such as in urban street-scenes, where the pixel-wise annotation process of a single image could span up to 90 minutes~\cite{cordts2016cityscapes}. This problem is amplified in the medical domain as we further require domain experts to annotate data who are severely time-restricted due to their clinical work.
Thus, we are faced with the problem of minimizing the needed annotation effort while maximizing model  accuracy.

The majority of existing approaches follow two orthogonal paths:
incorporating non-annotated data~\cite{hung2018adversarial} which is fast to obtain or using cheap weak labels with different levels of quality ranging from image-level \cite{chang2020weakly,  huang2018weakly,lee2019ficklenet,pathak2014fully} over single point \cite{bearman2016s} to bounding box annotations \cite{khoreva2017simple,song2019box}.
Both these so called semi- and weakly-supervised approaches led to great insights and convincing results.
However, they ignore that practical applications are often faced with several types of supervision simultaneously (\figref{fig:overview}).

In very common scenarios, we are provided with a small pixel-wise annotated data set, with automatically parsed image-level labels from medical reports, and with large amounts of additional unlabeled data from the same distribution.
Currently, it is largely unanswered how such diverse supervision types can be unified to train a semantic segmentation system.
To this end, we go beyond standard weakly- or semi-supervised learning and investigate also the semi-weakly supervised setting in the low-resource scenario of the medical domain. 

For dealing with only few annotated examples, we propose a novel pathway to integrate training signals deep into segmentation network layers via a new take on deep supervision.
We then amplify these signals by enriching weakly- or entirely un-labeled images via our novel approach to infer robust pseudo-labels using a mean-teacher segmentation model.

Furthermore, as iterative training processes in low-data scenarios are often unstable, we present experiments following a rigorous evaluation protocol and report test accuracies with standard deviations along numerous data-splits.

Our contributions amount to:
\begin{itemize}
    \item[(1)]We present the first thorough investigation of varying numbers of training samples and a large diversity of supervision types for semantic segmentation.
   \item[(2)]We introduce a novel perspective on the deep supervision paradigm adapting it to segmentation in our \emph{Multi-label Deep Supervision} technique.
   With this, we introduce a flexible semi-weakly supervised pathway to integrate either un- or weakly labeled images: our novel \emph{Self-Taught Deep Supervision} approach.
    \item[(3)] Finally, our best performing method \emph{Mean-Taught Deep Supervision} adds invariance towards perturbations and a robust pseudo-label generation, achieving results close to fully supervised baselines while using only a fraction of $5.78\%$ strong labels.
\end{itemize}

\section{Related Work}




\noindent\textbf{Mask supervision.}
The most prominent research direction in semantic segmentation considers the task in a fully-supervised setting, assuming the availability of a sufficient amount of expensive pixel-wise labeled masks~\cite{badrinarayanan2017segnet,chen2017deeplab,chen2018encoder, lin2017refinenet,long2015fully, ronneberger2015u, zhao2017pyramid}.
Based on this assumption, innovation in parameter-heavy models induces performance gains on large data sets~\cite{cordts2016cityscapes, everingham2015pascal, lin2014microsoft}, while in smaller-scale scenarios, \eg domains where label acquisition is even more expensive, leveraging this progress is only partially possible. 

\noindent\textbf{Bounding box supervision.}
Coarse bounding boxes drawn around semantic regions offer strong location cues at manageable costs, but inherently include pixels not semantically associated to the box-label.
On object-related data sets, leveraging this kind of supervision has led to impressive results~\cite{dai2015boxsup,khoreva2017simple, papandreou2015weakly,song2019box}.
Yet, methods tend to leverage hand-crafted rules or tools like GrabCut~\cite{rother2004grabcut},  Multiscale Combinatorial Grouping~\cite{arbelaez2014multiscale}, or Selective Search~\cite{uijlings2013selective} to kick-start training.
Therefore, such approaches are often not trivially transferred to data different from natural images.
In medical imaging, bounding box supervision for segmentation has been studied for positron emission tomography scans~\cite{afshari2019weakly} as well as magnetic resonance images~\cite{kervadec2020bounding, rajchl2016deepcut}.
In this work, we leverage box-supervision for retinal fluid segmentation in optical coherence tomography (OCT) scans. 

\noindent\textbf{Image-level supervision.}
Leveraging global labels with information about present classes in an image to train a segmentation model has recently seen growing interest in our community~\cite{chang2020weakly,  huang2018weakly,kolesnikov2016seed, lee2019ficklenet,pathak2014fully, wang2020self}.
While early approaches followed a Multiple Instance Learning (MIL) formulation~\cite{pathak2014fully}, recent work builds upon feature attribution methods~\cite{selvaraju2017grad,simonyan2014deep,zhou2016learning} for providing initial location cues and refine them using prior assumptions~\cite{chang2020weakly, huang2018weakly, kolesnikov2016seed, lee2019ficklenet, wang2020self}.
Such assumptions are often specific to the data set they were designed for, consequently having limited applicability for many medical domains.
In consequence, only few approaches are known which investigate benefits of image-level labels in context of OCT scans~\cite{schlegl2015predicting, yang2019weakly}.

\noindent\textbf{Semi- and semi-weakly supervision.}
Combining pixel-wise annotations with weaker annotations like global- or box-labels has been shown early in~\cite{papandreou2015weakly}. These scenarios are termed semi-weakly supervised segmentation~\cite{choe2020evaluating}.
With the emergence of generative adversarial networks, un- and pixel-wise labeled data were fused via adversarial objectives~\cite{hung2018adversarial,mittal2019semi,souly2017semi}, exploring semi-supervised segmentation models.
Making best use of bounding boxes in conjunction with pixel-wise labels has seen interest in~\cite{ibrahim2020semi,khoreva2017simple}.
Recently, many attempts try to regularize the learner to form predictions invariant towards perturbations~\cite{li2018semi,ouali2020semi}.
Along this line of thought,
the data augmentation strategy CutMix~\cite{yun2019cutmix} can serve as perturbation by mixing predictions and assembling new informative pseudo-labels~\cite{french2019semi,olsson2020classmix}.
This approach is often accompanied by mean-teachers~\cite{tarvainen2017mean} serving as robust models for distilling invariance into the student.
Closest to our work, medical image segmentation with a mean-teacher was done by changing the consistency term from~\cite{tarvainen2017mean} to better fit the segmentation task~\cite{perone2018deep}.
Our approach does not only enforce consistency between a teacher and student, we extend the idea with pseudo-targets.
Yet, instead of adding complexity by generating them iteratively~\cite{bai2017semi}, we integrate pseudo-labels online and identify the missing link in coping with their noise: deep supervision.

\noindent\textbf{Deep supervision.}
Early explorations of deep architectures for classification introduced companion objectives or dense pathways into lower layers~\cite{huang2017densely,lee2015deeply,szegedy2015going}.
The positive effects in low-data scenarios~\cite{yang2017towards}, on convergence, generalization and vanishing gradients~\cite{lee2015deeply} set off lots of work using deep supervision in segmentation models~\cite{chen2018semantic,liu2019dna,marmanis2016semantic}.
Especially the medical imaging community with their notoriously scarce-data applications showed strong  interest~\cite{abraham2019novel,dou20173d,fu2018joint,li2020deeply,xu2016gland,yang2017towards,zhang2018deep,zhou2017deep,zhou2018unet++}.
The move from classification to segmentation is mostly done by up-scaling low resolution feature maps.
We propose the counter intuitive reverse direction: down-scaling the pixel-wise annotations.
By doing so in a novel semantic-preserving fashion, we uncover surprising properties when joining it with noisy pseudo-labels.

\section{Proposed Approach}
\noindent In this chapter, we introduce notation and the task of \emph{semi-weakly supervised semantic segmentation}.
Thereafter, we present our novel perspective on deep supervision: \emph{Multi-label Deep Supervision} which we consider for traditional semantic segmentation and motivate its strong fit for semi- and semi-weakly supervised segmentation in our \emph{Self-Taught} and \emph{Mean-Taught Deep Supervision} frameworks.

\subsection{Preliminaries}

\subsubsection{Supervision modalities and notation}

For semi-weakly supervised semantic segmentation, we consider an image data set:
\begin{equation}
    \mathcal{D} = \{x_1, ..., x_n | x_i \in \mathbb{R}^{3 \times H \times W} \} \enspace.
    \label{eq:dataset}
\end{equation}
An image $x_i$ can have different available annotations, for example a pixel-wise annotated mask $m_i$, bounding box $b_i$, global image-level label $g_i$ from:
\begin{align}
    \label{eq:masks}
    \mathcal{M} &= \{m_1, ..., m_n| m_i \in [0,1]^{(C+1) \times H \times W}\} \enspace,\\
    \label{eq:bb}
    \mathcal{B} &= \{b_1, ..., b_n| b_i \in [0,1]^{(C+1) \times H \times W}\}\enspace,\\
    \label{eq:imagelevel}
    \mathcal{G} &= \{g_1, ..., g_n| g_i \in [0,1]^{(C+1)}\}\enspace,
\end{align}
or no annotation at all.
Note, for $m \in \mathcal{M}$, we assume that at each spatial position $(x,y) \in H \times W$ only a single class is set to one.
We replace the common representation of bounding boxes as two points $(x_1,y_1), (x_2,y_2) \in H \times W$ by a mask-like notation, where foreground classes $c$ are set to one for spatial positions falling inside their bounding box.
The background class $C+1$ is one at all left empty regions.

In semi-weakly supervised segmentation, we do not have full access on $\mathcal{M}$ for all images $x \in \mathcal{D}$ but each image is annotated with a mask, bounding box, image-level label or not at all.
Along with its special case of having masks and raw images, i.e. semi-supervised segmentation, the target is to correctly infer the semantic mask for any given image.

\subsubsection{Supervision integration}
\label{sec:supervision-integration}
Given an input feature map $f \in \mathbb{R}^{d \times H \times W}$, we refer to $H$ and $W$ as the spatial dimensions.
To form a prediction $p \in \mathbb{R}^{C \times H \times W}$ of the same spatial extent with $C$ output dimensions, in semantic segmentation, most commonly $f$ is transformed by $C$ $1\times1$ convolutions.
In our case, we transform $f$ by a sequence of $1\times1$ convolution, batch normalization~\cite{ioffe2015batch} and ReLU non-linearity followed by a final $1\times1$ convolution.
We refer to this as an output-head of the model.
In case we have multiple objective functions, we simply augment the model with additional output-heads on top of $f$, which is common practise in literature~\cite{bischke2019multi,teichmann2018multinet}.

Segmentation networks often employ encoder-decoder architectural designs, within which we later integrate auxiliary outputs (deep supervision).
More specifically, these additional outputs operate on feature maps $f_0, \dots, f_h$ produced in the \emph{decoding process}.
Here, we refer to $f_0$ as the innermost feature map in the decoder, while $f_h$ corresponds to the outermost feature map, on which standard segmentation models add an output-head for predictions.
Given the spatial dimensions $H_i \times W_i$ for the feature map $f_i$, we can generally assume that:
\begin{equation}
    \forall_{i \in \{0, ..., h\}; i < j}: H_i \leq H_j~\&~W_i \leq W_j \enspace.
    \label{eq:decoder_assumption}
\end{equation}
In our experiments, we employ encoder-decoder models for which $H_0 \ll H_h$ and $W_0 \ll W_h$ hold.
Computing predictions based on a certain feature map $f_i$ will be referred to as $\kappa_i(f_i) \in \mathbb{R}^{C \times H_i \times W_i}$, with $\kappa_i(\cdot)$ denoting an output-head.

\subsubsection{Supervision signals}
Different modes of supervision often require different loss functions.
The most common objective for training semantic segmentation models when pixel-wise masks are available is minimizing the \emph{cross-entropy loss}:
\begin{equation}
    \mathcal{L}_{CE}(f, m) = - \frac{1}{\Omega_{1}} \sum_{i,j,c=1}^{H,W,C}  m^{c,i,j} \cdot \log(\alpha(\kappa(f))^{c,i,j}) \,,
    \label{eq:cross_entropy}
\end{equation}
where $\alpha(\cdot)$ is the softmax along the first dimension and $\Omega_1 = H \cdot W$.
In case of multi-label settings, the \emph{binary cross-entropy loss} is commonly chosen:
\begin{align}
    BCE(o,t) &= t \cdot \log(\sigma(o)) + (1 - t)  \log(1 - \sigma(o))\\
    \mathcal{L}_{BCE}(f,m) &= - \frac{1}{\Omega_2} \sum_{i,j,c=1}^{H,W,C} BCE(\kappa(f)^{c,i,j}, m^{c,i,j})
    \label{eq:binarycrossentropy}
\end{align}
With $\Omega_2 = H \cdot W \cdot C$, this loss most commonly uses a sigmoid normalization $\sigma(\cdot)$.

\subsection{Multi-label Deeply Supervised Nets}
\begin{figure*}[t]
  \includegraphics[width=\textwidth]{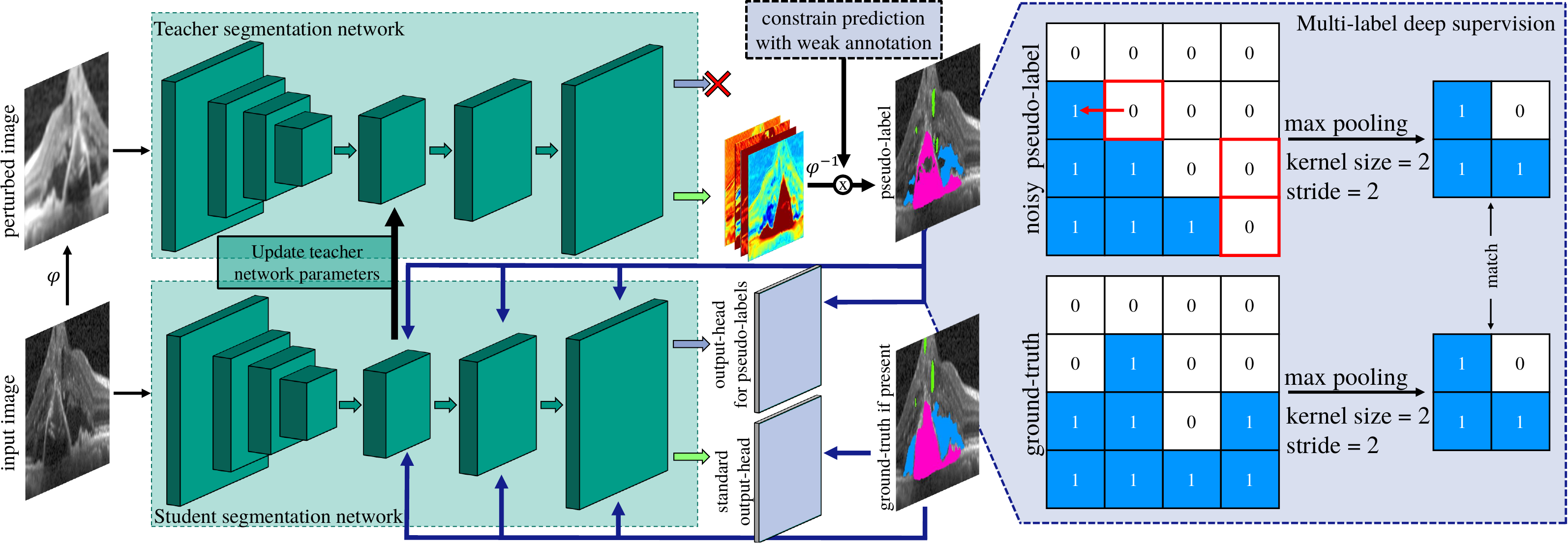}
  \caption{Our proposed approach combines pseudo-labeling via a mean-teacher with a novel perspective on deep supervision. By perturbing the input to the teacher and reversing geometric transformations in output-space, we streamline mean-teachers for segmentation. The new take on deep supervision, \ie \emph{Multi-label Deep Supervision} introduces a much needed smoothing effect for noisy pseudo-labels: At smaller scales, erroneous predictions (red) like small shifts or few missed pixel-classifications get smoothed out and match the unavailable mask.}
  \label{fig:method}
\end{figure*}


\noindent\textbf{Parameter-efficient multi-label deep supervision.}
In its inception, deep supervision was introduced for image classification~\cite{lee2015deeply}, as such, having to identify \emph{one} individual class corresponding to the input.
Yet, semantic segmentation in itself has a different target: find the class for \emph{each pixel} in the image.
The problem we identify is the way deep supervision is commonly integrated into segmentation networks.
Specifically, the challenge arises due to the mismatch in spatial dimensions between the full-scale ground-truth mask and the smaller spatial resolution within the network's feature maps (Equation~\ref{eq:decoder_assumption}).

Apart from~\cite{liu2020mdan} which use a lossy nearest interpolation, most work forces the network to learn an up-scaling to address the mismatch between low-resolution spatial features and the ground-truth mask.
Then, after up-scaling, a standard output-head transforms the features into a prediction with the same size as the mask.
Consider the toy example with the feature map $f_{small} \in \mathbb{R}^{d\times 10\times 10}$ and a corresponding ground-truth mask $m_{big} \in \mathbb{R}^{c\times 100 \times 100}$.
To enrich the segmentation model via deep supervision, previous approaches would for the $d$ dimensional feature $f^{:,x,y}_{small}$ at the spatial location $(x,y) \in H_{small}\times W_{small}$ up-scale and infer a patch of size $10 \times 10$ in the ground-truth.
Not only is this a tremendously hard task, it is precisely the task the entirety of the decoder is trying to solve.

To summarize, we identify two shortcomings, (1) the network has to learn an up-scaling, at the cost of additional parameters and (2) intermediate features are burdened to model complex classification information and spatial relations in \emph{output space} that we question to be useful in the decoding process, but presumably only serve as skip-connections for more stable gradients.

Instead, we propose to model each feature vector $f^{:,x,y}$ at each location $(x,y)$ in feature maps as patch-descriptors for their receptive field in the input image.
Thus, we aim at ingraining the semantic information of all pixels contained in the receptive field of the patch-descriptors into the model.
We argue that this can be achieved by enforcing a multi-label loss with a label containing \emph{all semantic classes} present in the receptive field.
It follows, that instead of up-scaling feature maps, we can simply down-scale the ground-truth mask to match the size of the feature maps at no cost of parameters and contain labels of \emph{all classes confined} in the receptive fields to preserve semantic information.

By formulating ground-truth masks as binary tensors~\eqref{eq:masks}, this down-scaling process can be efficiently implemented, by applying max-pooling with a fitting kernel-size and stride to match the feature map's spatial extent.
The new down-scaled target~$m^*_i \in \mathbb{R}^{C\times H_i \times W_i}$, therefore, contains the multi-label ground-truth $m^{*~:,x,y}_i$ of the feature $f^{:,x,y}_i$ (i.e. patch-descriptor) aggregated from all spatial positions within the feature's receptive field.

\begin{figure}[t]
  \includegraphics[width=\linewidth]{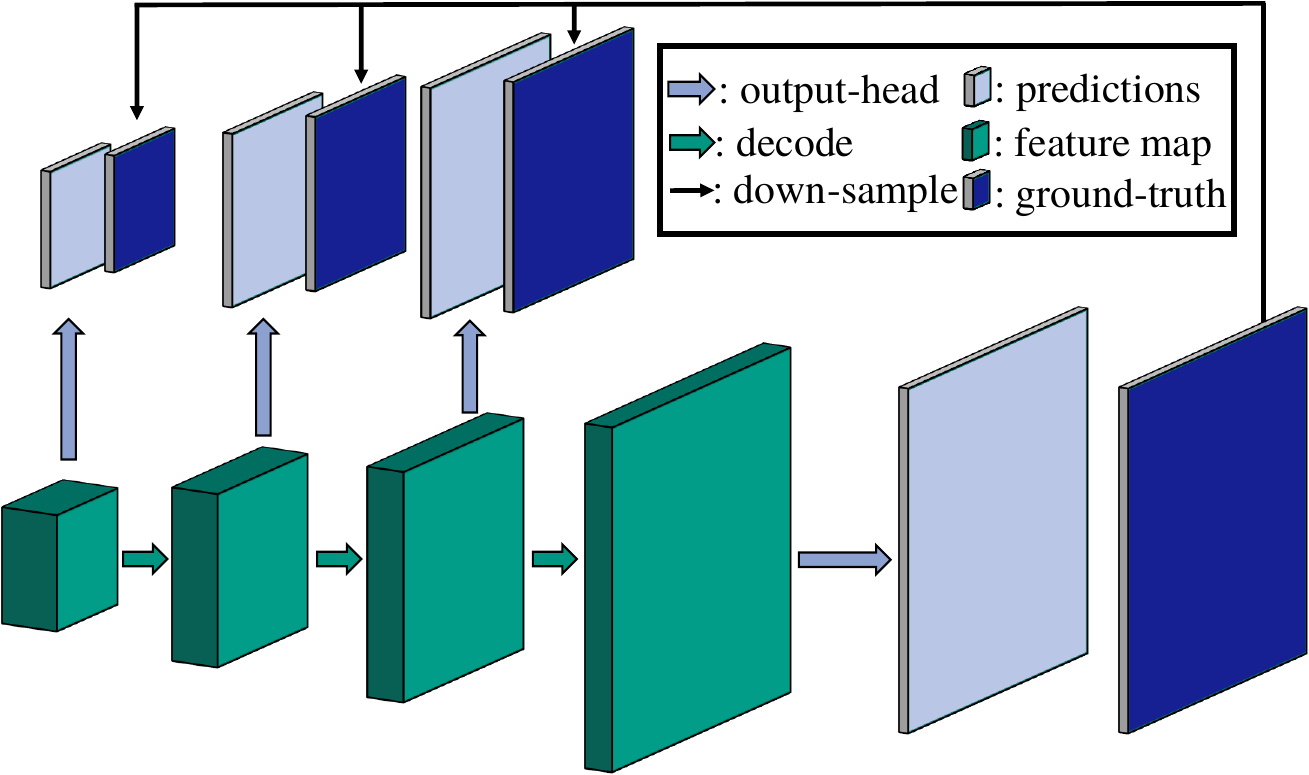}
  \caption{Our proposed method of integrating deep supervision into the decoder of segmentation networks by down-sampling the pixel-mask and enforcing a multi-label classification loss.}
  \label{fig:multi-label-deep-supervision}
\end{figure}

As shown in~\figref{fig:multi-label-deep-supervision}, we integrate such semantic-preserving, down-scaled multi-label ground-truths by applying output-heads on top of the feature maps and enforce:
\begin{equation}
    \mathcal{L}(f_1,\dots,f_h,m^*_1,\dots,m^*_h) = \frac{1}{h} \sum_{k=1}^{h} \mathcal{L}_{BCE}(f_k, m^*_k).
    \label{eq:mlds_loss}
\end{equation}
We refer to this way of integrating supervision as \emph{Multi-label Deep Supervision} since it respects the presence of multiple classes within the receptive field at each spatial location in the hierarchical, low-resolution feature maps.
During training, we pair this loss for the outermost feature maps with a \emph{standard output-head} using~\eqref{eq:cross_entropy}, and only use the latter for inference in a usual fashion.

\noindent\textbf{Self-taught deep supervision.}
\label{sec:self-taught-deep-supervison}
For semi-supervised segmentation, some authors consider generating pseudo-labels for unlabeled images to add noisy annotated samples.
In a similar fashion, we propose to generate pseudo-labels but integrate them by providing \emph{Multi-label Deep Supervision}.
The motivation behind this is that pseudo-labels at full resolution will contain a significant amount of incorrect predictions.
Yet, making use of \emph{Multi-label Deep Supervision} we down-sample the pseudo-labels, naturally smoothing the noisy supervision signal.
For earlier decoder outputs with smaller spatial resolution, the down-sampled pseudo-label will better match the unavailable actual ground-truth, as shown on the right-hand side of~\figref{fig:method}.

We introduce \emph{Self-Taught Deep Supervision}, which generates a binary ground-truth tensor for an unlabeled sample in the following fashion: First the unlabeled image $x_i$ is passed through the segmentation network to obtain a prediction $p_i$, which we transform into a pseudo-mask by attributing ones to classes with the highest probability at each location $(x,y)$.
Using this pseudo-label, as described before, we enforce \emph{Multi-label Deep Supervision}.
We use separate output-heads on the outermost feature map for the ground-truth and the generated pseudo-labels, which turned out to be important.
Thus, we integrate one output-head which is trained with standard cross-entropy loss and the clean labels, as well as a second output-head which is trained as part of Eq.~\eqref{eq:mlds_loss} using pseudo-labels.
For inferring pseudo-labels, we use the output-head trained with clean labels (see~\figref{fig:method}).

In case an image-level label $g_i$ is available, we can further constrain the generated pseudo-label to the classes contained in $g_i$.
In a similar fashion, an associated bounding box label $b_i$ can constrain the pseudo-labels to lie within coarse regions.
This results in a flexible integration of weakly-labeled images to improve pseudo-label quality.

\noindent\textbf{Mean-taught deep supervision.}
In our final variant, we extend the \emph{Self-Taught Deep Supervision} approach with a stronger pseudo-label generation.
We generate more robust pseudo-labels by (1) enforcing consistent predictions with respect to perturbations and (2) using a teacher model, which is a combination of all models of previous iterations.

Mean-teachers (i.e. the exponential moving average over previous model parameters) were successfully introduced for semi-supervised classification~\cite{tarvainen2017mean} and saw some use in segmentation~\cite{french2019semi,olsson2020classmix,perone2018deep}.
The idea is that better predictions can be obtained by maintaining a teacher model via continuously updating its parameters with the moving average of the student model and the previous teacher model.
Hence, the teacher is not trained separately, but just updated by:
\begin{equation}
    \theta_{t}^{\text{teacher}} = \alpha \cdot \theta_{t-1}^{\text{teacher}} + (1 - \alpha )\cdot \theta_{t}^{\text{student}} \enspace.
    \label{eq:mean_teacher}
\end{equation}
Here, $\theta$ denotes either the parameters of the student- or teacher model, while the index denotes the training iteration and $\alpha$ the smoothing coefficient.
The authors of~\cite{tarvainen2017mean} use the teacher to align its softmax predictions with the student's given differently perturbed inputs using a mean squared error (MSE) loss.
We extend this by leveraging the mean-teacher to provide pseudo-labels for our \emph{Multi-label Deep Supervision} that the student network is trained with.

Previous adaptations to semantic segmentation either did not perturb the input image to the teacher~\cite{perone2018deep} or used CutMix variants~\cite{french2019semi, olsson2020classmix} as perturbations.
Instead, we propose to perturb the input image to the teacher in a different way as the student input and still obtain matching outputs.
To this intent, we transform the teacher's input $x_i$ by photo-metric (\eg color jittering) and geometric perturbations (\eg flipping) $\varphi(x_i)$. After the forward pass, we reverse only the geometric perturbations on the predictions via~$\varphi^{-1}(p_i)$. 
This alteration is the crucial detail to make mean-teachers work for semantic segmentation using common, simple data augmentations as perturbations.
As the mean-teacher is deemed to produce better predictions, we use its \emph{standard output-head} for inference.
See also~\figref{fig:method} for visualizations.


\section{Experiments}


\subsection{Data set}
\noindent Our experiments are built on the publicly available RETOUCH data set~\cite{RETOUCH} for retinal fluid segmentation.
It is a collection of optical coherence tomography (OCT) volumes (stacks of b-scans) containing different retinal diseases.
These volumes can be obtained by imaging tools of different vendors of which the data set features three: Spectralis, Cirrus, and Topcon.
Generally, b-scans differ across manufacturers in appearance and will be considered separately in our experiments.
Further, as part of our work lies in investigating the scarce-data scenario, our main experiments are carried out on the smallest of the three, Spectralis ($49$ b-scans per volume), while the performance on the remaining vendors ($128$ b-scans per volume) is evaluated to emphasize the generalization of our approach.
The data set is fully annotated with pixel-wise labels for three types of retinal fluids: Intraretinal fluid, Subretinal fluid and Pigment Epithelial Detachments.
For our experiments, we derive bounding boxes and image-level labels from the masks.

\noindent\textbf{Experimental setup.}
As low-data scenarios entail sensitivity in the optimization process, we carry out each experimental configuration on $10$ cross-validation splits.
Each split is generated randomly and independent for each vendor: First, we randomly select $5$ volumes for validation and $5$ for testing.
The remaining volumes (Spectralis: $14$, Cirrus: $14$, Topcon: $12$) containing fluid annotations form the training set.
In our experiments, we consider scenarios where $3,6,12,24$ labeled b-scans are present (corresponding to $1,2,4,8$ b-scans \emph{per class}), therefore, we enumerate the scans in the training set and make sure that in an interval of size $3$ all diseases are present (as long as available).
Thus, \eg the first $6$ images in a split contain each disease at least twice, which ensures that all models access the exact same annotations and the scenarios with more annotations subsume the smaller ones (\eg $\mathcal{D}_{3} \subset \ldots \subset \mathcal{D}_{24}$).

We consider two settings: (1) the models are provided with pixel-wise labeled masks ($\mathcal{M}$) or (2) with coarse bounding boxes ($\mathcal{B}$) as strongest annotation.
In those scenarios, we train models having access to either one of these labels ($\mathcal{M}$ or $\mathcal{B}$), models having additionally access to the remaining unlabeled examples ($\mathcal{U}$, i.e. semi-supervised) or having access to the remaining b-scans with global image-level labels ($\mathcal{G}$, i.e. semi-weakly supervised).

\noindent\textbf{Evaluation metric.}
Our evaluation follows the standard procedure for segmentation models.
We calculate the prediction for all pixels in all testing images and then calculate the Intersection over Union (IoU) for all disease classes with the corresponding ground-truth pixels.
Then, these disease-wise IoUs are averaged resulting in the mean IoU (mIoU).
We carry out each experiment on all splits ($10$ trained models for one result), the final measures we report are the average mIoU and standard deviation over the splits.

\subsection{Implementation details}
\label{sec:implementation-details}
\noindent\textbf{Pre-processing.}
We employ the pre-processing of~\cite{apostolopoulos2017retinet} on the individual b-scans which includes anisotropic filtering and warping the lower retinal edge onto a straight line.
Finally, we resize all scans to the size $200\times200$.

\noindent\textbf{Network parameters.}
For our encoder-decoder segmentation model, we use a standard UNet architecture~\cite{ronneberger2015u}, ubiquitous in medical image analysis.
The UNet is augmented with batchnorm layers~\cite{ioffe2015batch} and a total of four down-scaling and four up-scaling (bilinear interpolation) convolutional blocks.
This way, the decoder has a total of four feature maps $f_0,\dots,f_3$ increasing in spatial dimensions as well as a feature map $f_4$ directly before the standard output $1\times1$ convolution.
As noted before, we exchange the output layers with output-heads as stated in~\secref{sec:supervision-integration}, enabling each head to learn an individual non-linear transformation of the features.
In case we use deep supervision, we add output-heads for all feature maps $f_{\{0-4\}}$ and enforce the respective loss function.
All experiments are carried out with the same basic training-hyperparameters: 
We use a batch size of $16$ b-scans, which are transformed by a horizontal flip $50\%$ of the time and by a random adjustment of brightness, contrast, hue and saturation by a factor of $0.0-0.1$.
Network parameters are initialized using Xavier initialization~\cite{glorot2010understanding} and the optimization process spans over $100$ epochs with SGD optimization and a momentum term of $0.9$. 
The learningrate is adjusted once after 80 epochs from $0.01$ to $0.001$.
Each model is evaluated every 10 epochs on the validation set, we perform early stopping and at the end of our experimentation phase evaluate the best model once on the testing set.

\begin{table}[t]
\centering
\begin{tabular}{c|c|c|c|c}
    larger $\varphi$ & inference & $\alpha$ & MSE & validation (mIoU)\\
    \hline
    \checkmark & teacher & $0.5$ & -- & $61.24 \pm 3.69$\\
    \checkmark & teacher & $0.5$ & \checkmark & $\textbf{61.36} \pm \textbf{4.73}$\\
    \checkmark & teacher & $0.1$ & \checkmark & $60.15 \pm 4.14$\\
    \cdashline{1-5}
    \checkmark & student & $0.1$ & \checkmark & $58.54 \pm 3.62$\\
    -- & student & $0.1$ & \checkmark & $58.26 \pm 4.27$\\
    \hline
    \hline
    -- & student & $0.0$ & -- & $57.80 \pm 4.68$ \\
    \hline
\end{tabular}
\caption{Ablation for \emph{Mean-Taught Deep Supervision} using $24$ pixel-masks and the remaining image-level labels. Last line shows \emph{Self-Taught Deep Supervision} performance for comparison.}
\label{fig:mean_teacher_ablation}
\end{table}

\subsection{Baselines and methods}
\noindent Our first \emph{Baseline} model is a UNet only using the strongest available supervision (mask or bounding box) as training signal, integrating it via the loss of~\eqref{eq:cross_entropy} on $f_4$.
To accommodate for the small amount of seen images (only $\{3,6,12,24\}$), we train the model for $10$ times the epochs.
All other models should outperform this model.

\noindent Next, we consider the semi-supervised scenario. Here, we leverage a UNet which uses~\eqref{eq:cross_entropy} for available masks or bounding boxes and additionally the self-supervised invariant information clustering (IIC) loss of~\cite{ji2019invariant} for all labeled and unlabeled images on $f_4$.
This \emph{IIC Baseline} sets the bar for all models integrating unlabeled data.

\setlength{\tabcolsep}{2.5pt}
\begin{table*}[t]
\centering
\begin{tabular}{r|cc|c|c|c|c||c|}
    Method & $\mathcal{G}$ & $\mathcal{U}$ & $3$ & $6$ & $12$ & $24$ & Full Access \\
    \hline
    &&&\multicolumn{5}{c|}{\textbf{Mask Supervision}}\\
    \hline
    Baseline~\cite{ronneberger2015u} &  &  & $14.80 \pm \phantom{0}6.50$ & $26.98 \pm \phantom{0}7.83$ & $35.39 \pm 6.36$ & $48.63 \pm 5.17$ & $62.09 \pm 4.77$\\
    Multi-Label Deep Supervision (Ours) &  &  & $\textbf{17.98} \pm \phantom{0}\textbf{8.20}$ & $\textbf{32.92} \pm \phantom{0}\textbf{7.35}$ & $\textbf{42.96} \pm \textbf{6.71}$ & $\textbf{52.68} \pm \textbf{6.82}$ & $\textbf{65.82} \pm \textbf{4.64}$\\
    \cdashline{1-8}
    IIC Baseline$^{8}$~\cite{ji2019invariant} &  & \checkmark & $\textbf{22.45} \pm \phantom{0}\textbf{9.36}$ & $32.02 \pm \phantom{0}7.23$ & $41.48 \pm 7.26$ & $53.08 \pm 6.13$ & $65.16 \pm 3.80$\\
    Deeply Supervised IIC$^{8}$ &  & \checkmark & $20.78 \pm \phantom{0}8.83$ & $31.39 \pm 10.26$ & $39.18 \pm 6.94$ & $50.10 \pm 7.92$ & $65.18 \pm 3.85$\\
    \emph{Perone and Cohen-Adad}$^{10}$~\cite{perone2018deep} &  & \checkmark & $16.17 \pm 10.74$ & $33.10 \pm 10.24$ & $45.80 \pm 7.51$ & $54.75 \pm 5.96$ & $65.49 \pm 4.14$\\
    Self-Taught Deep Supervision (Ours) &  & \checkmark & $10.37 \pm \phantom{0}8.29$ & $28.62 \pm 12.96$ & $43.57 \pm 9.97$ & $56.11 \pm 6.30$ & $66.24 \pm 4.67$\\
    Mean-Taught Deep Supervision$^{10}$ (Ours) &  & \checkmark & $16.31 \pm 15.48$ & $\textbf{35.17} \pm \textbf{11.35}$ & $\textbf{53.52} \pm \textbf{8.72}$ & $\textbf{58.84} \pm \textbf{6.57}$ & $\textbf{66.31} \pm \textbf{4.66}$\\
    \cdashline{1-8}
    MIL Baseline & \checkmark &  & $15.44 \pm 11.10$ & $25.46 \pm \phantom{0}8.57$ & $41.34 \pm 9.66$ & $49.07 \pm 8.20$ & $61.50 \pm 5.64$\\
    Deeply Supervised MIL & \checkmark &  & $20.02 \pm \phantom{0}9.17$ & $31.50 \pm \phantom{0}8.88$ & $44.29 \pm 5.03$ & $51.13 \pm 3.93$ & $62.04 \pm 3.92$\\
    Self-Taught Deep Supervision (Ours) & \checkmark &  & $20.47 \pm \phantom{0}8.62$ & $36.40 \pm \phantom{0}8.91$ & $49.39 \pm 9.95$ & $59.29 \pm 7.52$ & $66.34 \pm 3.81$\\
    Mean-Taught Deep Supervision$^{10}$ (Ours) & \checkmark &  & $\textbf{21.91} \pm \textbf{13.49}$ & $\textbf{42.14} \pm \textbf{14.25}$ & $\textbf{54.70} \pm \textbf{9.26}$   & $\textbf{60.45} \pm \textbf{5.71}$ & $\textbf{66.39} \pm \textbf{4.29}$\\
    \hline
    &&&\multicolumn{5}{c|}{\textbf{Bounding Box Supervision}}\\
    \hline
    Baseline~\cite{ronneberger2015u} &  &  & $12.49 \pm 4.28$ & $18.32 \pm 4.94$ & $25.62 \pm 3.08$ & $29.55 \pm 2.77$ & $38.45 \pm 4.44$\\
    Multi-Label Deep Supervision (Ours) &  &  & $\textbf{14.59} \pm \textbf{5.81}$ & $\textbf{19.62} \pm \textbf{6.21}$ & $\textbf{27.89} \pm \textbf{3.44}$ & $\textbf{32.02} \pm \textbf{4.78}$ & $\textbf{38.66} \pm \textbf{3.36}$\\
    \cdashline{1-8}
    IIC Baseline$^{8}$~\cite{ji2019invariant} &  & \checkmark & $\textbf{15.40} \pm \textbf{7.07}$ & $18.15 \pm 7.49$ & $26.05 \pm 6.00$ & $30.07 \pm 4.32$ & $38.45 \pm 4.65$\\
    Deeply Supervised IIC$^{8}$ &  & \checkmark & $12.77 \pm 7.15$ & $17.76 \pm 6.26$ & $\textbf{28.99} \pm \textbf{4.60}$ & $30.64 \pm 3.05$ & $38.81 \pm 4.48$\\
    \emph{Perone and Cohen-Adad}$^{10}$~\cite{perone2018deep} &  & \checkmark & $11.17 \pm 7.41$ & $\textbf{19.02} \pm \textbf{8.46}$ & $27.44 \pm 5.81$ & $31.72 \pm 3.87$ & $39.38 \pm 3.56$\\
    Self-Taught Deep Supervision (Ours) &  & \checkmark & $\phantom{0}5.14 \pm 3.84$ & $\phantom{0}9.62 \pm 7.35$ & $24.47 \pm 6.12$ & $32.71 \pm 3.56$ & $\textbf{39.39} \pm \textbf{3.63}$\\
    Mean-Taught Deep Supervision$^{10}$ (Ours) &  & \checkmark & $\phantom{0}8.21 \pm 3.96$ & $14.28 \pm 7.48$ & $24.79 \pm 5.79$ & $\textbf{34.14} \pm \textbf{3.10}$ & $39.04 \pm 4.15$\\
    \cdashline{1-8}
    MIL Baseline & \checkmark &  & $15.82 \pm 6.55$ & $16.95 \pm 6.19$ & $22.56 \pm 4.56$ & $26.48 \pm 5.51$ & $37.15 \pm 4.06$\\
    Deeply Supervised MIL & \checkmark &  & $\textbf{17.14} \pm \textbf{8.06}$ & $20.18 \pm 4.61$ & $24.15 \pm 4.95$ & $29.12 \pm 4.75$ & $37.94 \pm 3.35$\\
    Self-Taught Deep Supervision (Ours) & \checkmark &  & $16.04 \pm 8.52$ & $\textbf{22.15} \pm \textbf{6.29}$ & $28.63 \pm 4.04$ & $32.37 \pm 3.75$ & $\textbf{38.97} \pm \textbf{3.59}$\\
    Mean-Taught Deep Supervision$^{10}$ (Ours) & \checkmark &  & $15.81 \pm 8.59$ & $21.97 \pm 8.17$ & $\textbf{29.83} \pm \textbf{5.30}$ & $\textbf{34.81} \pm \textbf{3.62}$ & $38.66 \pm 4.73$\\
    \hline
\end{tabular}
\caption{Results for a diverse variety of models using varying amounts ($3$,$6$,$12$,$24$,all) of annotations, i.e. ground-truth masks or coarse bounding boxes. Approaches with $\mathcal{U}$ use additional unlabeled images, with $\mathcal{G}$ global image labels (\textbf{best results in category bold}). Accuracy in average mIoU over 10 testing splits with standard deviation. Superscripts indicate smaller batch sizes due to memory constraints.}
\label{tab:mask_and_boundingbox}
\end{table*}

\noindent For the semi-weakly supervised scenario with access to image-level labels $g$, we introduce a Multiple-Instance Learning (MIL) model.
This lower bound leverages the weakly labeled images by pooling a feature map in the spatial dimension, classifying this pooled feature $\bar{f}$ and enforcing $\mathcal{L}_{BCE}(\bar{f}, g)$ from~\eqref{eq:binarycrossentropy}.
Here, we use a pixel-wise cross-entropy loss for either masks or bounding boxes on $f_4$ and for image-level labeled images we pool $f_4$ via average pooling enforcing the MIL loss.

\noindent To show the effect of integrating deep supervision into segmentation models, we show performance of the previous two models but use all feature maps $f_{\{0-4\}}$.
We term these models \emph{Deeply Supervised IIC} and \emph{Deeply Supervised MIL}.
As we will see, the latter will provide an even stronger lower bound on semi-weakly supervised segmentation.

\noindent To grasp whether \emph{Multi-Label Deep Supervision} is beneficial in training, we extend the \emph{Baseline} UNet with our novel loss integration for all decoder feature maps $f_{\{0-4\}}$.

\noindent\emph{Self-Taught Deep Supervision} can be applied to both the semi- and semi-weakly segmentation scenario. We integrate pseudo-labels as described in~\secref{sec:self-taught-deep-supervison}. When image-level labels are available, we refine the noisy pseudo-labels by zeroing out absent classes.
As a natural extension of this approach, \emph{Mean-Taught Deep Supervision} is setup similarly.

\noindent Further, we benchmark the consistency-based approach of \emph{Perone and Cohen-Adad}~\cite{perone2018deep}.
Re-implementing their approach as described led to diverging models.
Thus, we modify it by using cross-entropy- instead of the DICE loss~\cite{milletari2016v}, $\alpha=0.5$, no ramp up phase and balanced loss weighting.

\subsection{Ablation studies}
\noindent For \emph{Mean-Taught Deep Supervision}, we perform a sequence of ablation experiments on the validation set as shown in Table~\ref{fig:mean_teacher_ablation}.
First, we move from not using a teacher model (\emph{Self-Taught Deep Supervision}) to using one.
Further improvement is made when adding stronger perturbations $\varphi(\cdot)$. To this extent instead of using a factor $0.1$, we use $0.4$ for the photo-metric components described in~\secref{sec:implementation-details}.
Inferring predictions with the teacher instead of the student for inference and tuning the hyperparameter $\alpha$ led to the best configuration.
To show that the mean squared error term taken over from~\cite{tarvainen2017mean} is not the vital component, we omit it and observe a minor decrease of $0.12\%$ mIoU.

\setlength{\tabcolsep}{4.5pt}


\begin{table*}[t]
\centering
\begin{tabular}{r|c|c|c|c||c|}
    Method & $\mathcal{U}$ & $6$ & $12$ & $24$ & Full Access\\
    \hline
    &&\multicolumn{4}{c|}{\textbf{Cirrus}}\\
    \hline
    Baseline~\cite{ronneberger2015u} &  & $12.31 \pm 5.41$ & $19.43 \pm 8.00$ & $30.10 \pm \phantom{0}9.34$ & $48.92 \pm 11.94$\\
    Multi-label Deep Supervision (Ours) &  & $\textbf{15.99} \pm \textbf{6.87}$ & $\textbf{25.12} \pm \textbf{8.58}$ & $33.53 \pm \phantom{0}9.44$ & $50.47 \pm 10.84$\\
    \emph{Perone and Cohen-Adad}$^{10}$~\cite{perone2018deep} & \checkmark & $12.36 \pm 6.12$ & $24.99 \pm 6.49$ & $33.79 \pm 10.15$ & $49.75 \pm 12.87$\\
    Mean-Taught Deep Supervision$^{10}$ (Ours) & \checkmark & $\phantom{0}9.18 \pm 8.53$ & $23.33 \pm 7.37$ & $\textbf{35.82} \pm \textbf{11.40}$ & $\textbf{51.24} \pm \textbf{10.94}$\\
    \hline
    &&\multicolumn{4}{c|}{\textbf{Topcon}}\\
    \hline
    Baseline~\cite{ronneberger2015u} &  & $14.79 \pm \phantom{0}9.34$ & $21.19 \pm 11.57$ & $27.61 \pm 10.31$ & $42.22 \pm 10.42$\\
    Multi-label Deep Supervision (Ours) &  & $\textbf{18.20} \pm \textbf{10.48}$ & $20.92 \pm 13.02$ & $33.71 \pm 11.92$ & $\textbf{45.85} \pm \textbf{10.32}$\\
    \emph{Perone and Cohen-Adad}$^{10}$~\cite{perone2018deep} & \checkmark & $15.26 \pm 12.74$ & $21.88 \pm 12.48$ & $27.67 \pm 13.81$ & $41.43 \pm \phantom{0}8.18$\\
    Mean-Taught Deep Supervision$^{10}$ (Ours) & \checkmark & $14.39 \pm 11.19$ & $\textbf{23.92} \pm \textbf{15.25}$ & $\textbf{33.87} \pm \phantom{0}\textbf{8.25}$ & $42.70 \pm 10.97$\\
    \hline
\end{tabular}
\caption{Segmentation accuracy of different methods on Cirrus and Topcon OCT vendors in average mIoU over $10$ testing splits.}
\label{tab:vendors}
\end{table*}

\subsection{Quantitative results}
\noindent\textbf{Mask supervision.}
In the upper half of Table~\ref{tab:mask_and_boundingbox}, we study results of the diverse methods using pixel-wise annotations.
The first observation we can make is that just one sample per class, i.e. the $3$-scenario, is not sufficient as all methods struggle in producing meaningful segmentations.
The step to $2$ pixel-wise annotations per class sounds small, yet we can observe a giant leap of $20.23\%$ to respectable $42.14\%$ mIoU for the \emph{Mean-Taught Deep Supervision} model which integrates weak global labels.
Throughout the scenarios $6$, $12$, and $24$ annotated samples, our \emph{Mean-Taught Deep Supervision} outperforms all other models consistently in both semi- and semi-weakly supervised segmentation.
With just $24$ masks (\ie $5.78\%$ of all annotations), our results come close to the full access \emph{Baseline} that requires $415.5$ masks (average over splits).
Note the large margins between \emph{Perone and Cohen-Adad} and our mean-teacher semi-supervised model, stretching up to $8.12\%$.
Integrating our \emph{Multi-label Deep Supervision} as compared to the standard Cross-Entropy loss (\emph{Baseline}) increases the performance by $3.18\%$, $5.94\%$, $7.57\%$, $4.05\%$ and even fully supervised by $3.73\%$ showcasing the strength of our novel deep supervision mechanism.

{\renewcommand{\arraystretch}{0.35} 
\setlength{\abovecaptionskip}{2pt}
\setlength{\belowcaptionskip}{-15pt}
\setlength{\tabcolsep}{1.5pt}

\begin{figure*}
    \centering
    \begin{tabular}{|cccccc|}
    \hline
        \multicolumn{6}{|c|}{}\\
            Input &
            $6$&
            $12$&
            $24$&
            Full Access&
            Target

            \\[0.5ex]
            \hline
          \multicolumn{6}{|c|}{}\\[0.1ex]
          
          \frame{\includegraphics[width=0.15\linewidth]{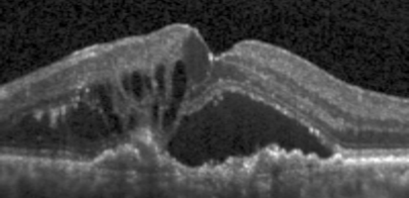}} & 
          \frame{\includegraphics[width=0.15\linewidth]{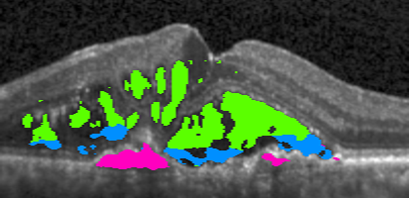}} & 
          \frame{\includegraphics[width=0.15\linewidth]{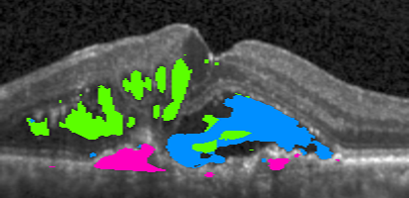}} & 
          \frame{\includegraphics[width=0.15\linewidth]{latex/figures/evaluation_figures/images/baseline/012_prediction.png}} & 
          \frame{\includegraphics[width=0.15\linewidth]{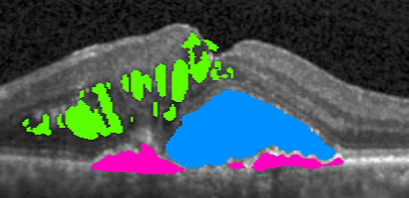}} & 
          \frame{\includegraphics[width=0.15\linewidth]{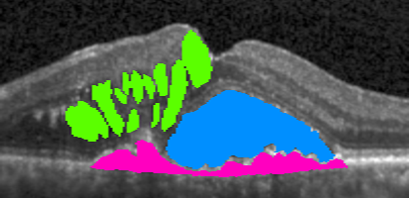}} \\
          \multicolumn{6}{|c|}{Baseline}\\[0.5ex]

          
          \frame{\includegraphics[width=0.15\linewidth]{latex/figures/evaluation_figures/images/others/input.png}} & 
          \frame{\includegraphics[width=0.15\linewidth]{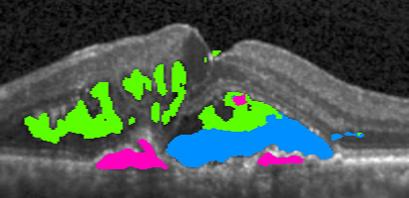}} & 
          \frame{\includegraphics[width=0.15\linewidth]{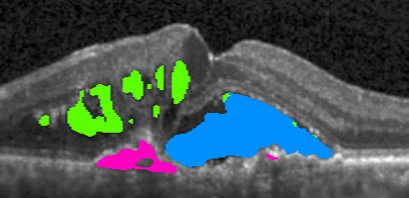}} & 
       \frame{\includegraphics[width=0.15\linewidth]{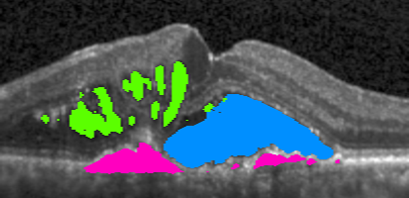}} & 
          \frame{\includegraphics[width=0.15\linewidth]{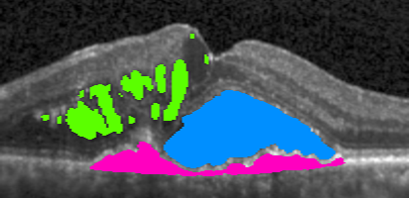}} & 
          \frame{\includegraphics[width=0.15\linewidth]{latex/figures/evaluation_figures/images/others/target.png}} \\
          \multicolumn{6}{|c|}{\textbf{Our approach} -- Multi-label Deep Supervision}\\[0.3ex]
          \hline\multicolumn{6}{|c|}{}\\

          
          \frame{\includegraphics[width=0.15\linewidth]{latex/figures/evaluation_figures/images/others/input.png}} & 
          \frame{\includegraphics[width=0.15\linewidth]{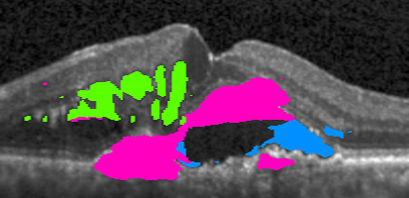}} & 
          \frame{\includegraphics[width=0.15\linewidth]{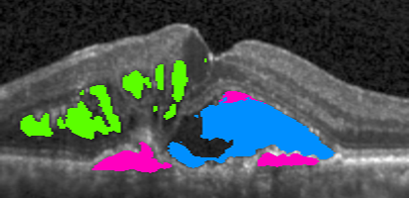}} & 
          \frame{\includegraphics[width=0.15\linewidth]{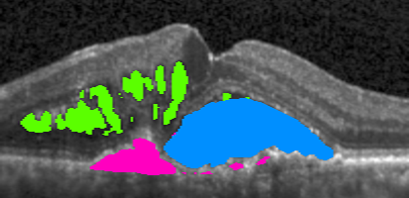}} & 
          \frame{\includegraphics[width=0.15\linewidth]{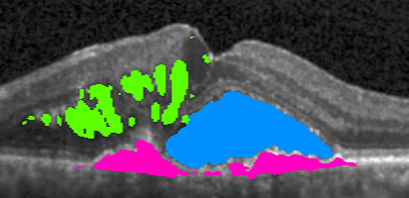}} & 
          \frame{\includegraphics[width=0.15\linewidth]{latex/figures/evaluation_figures/images/others/target.png}} \\
         \multicolumn{6}{|c|}{MIL Baseline}\\[0.5ex]

          
          \frame{\includegraphics[width=0.15\linewidth]{latex/figures/evaluation_figures/images/others/input.png}} & 
          \frame{\includegraphics[width=0.15\linewidth]{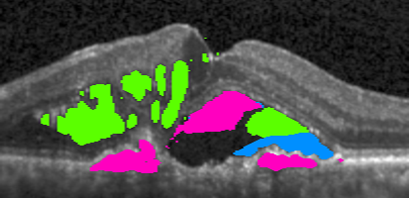}} & 
          \frame{\includegraphics[width=0.15\linewidth]{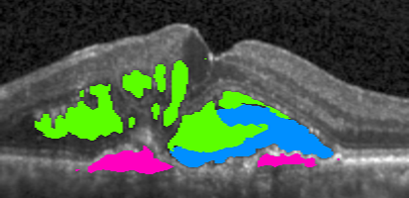}} & 
          \frame{\includegraphics[width=0.15\linewidth]{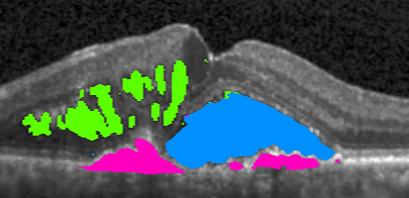}} & 
          \frame{\includegraphics[width=0.15\linewidth]{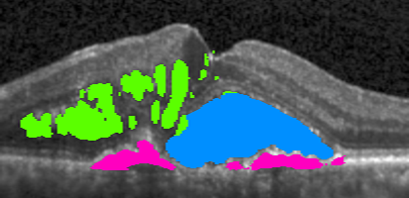}} & 
          \frame{\includegraphics[width=0.15\linewidth]{latex/figures/evaluation_figures/images/others/target.png}} \\
          \multicolumn{6}{|c|}{Deeply Supervised MIL}\\[0.5ex]


          \frame{\includegraphics[width=0.15\linewidth]{latex/figures/evaluation_figures/images/others/input.png}} & 
          \frame{\includegraphics[width=0.15\linewidth]{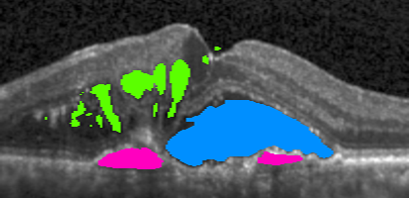}} & 
          \frame{\includegraphics[width=0.15\linewidth]{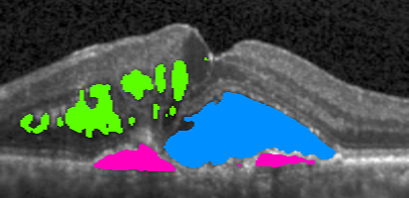}} & 
          \frame{\includegraphics[width=0.15\linewidth]{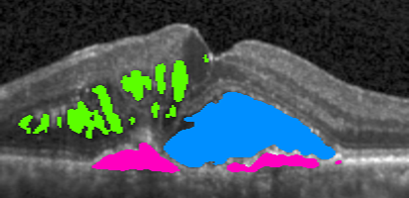}} & 
          \frame{\includegraphics[width=0.15\linewidth]{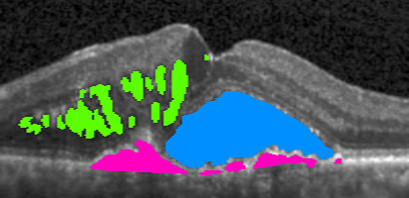}} & 
          \frame{\includegraphics[width=0.15\linewidth]{latex/figures/evaluation_figures/images/others/target.png}} \\
          \multicolumn{6}{|c|}{\textbf{Our approach} -- Mean-Taught Deep Supervision}\\[0.5ex]
          \hline
         
    \end{tabular}
    \caption{Segmentation progression when increasing the number of pixel-wise annotated masks from $6$ to $12$, $24$, and full access. The upper two methods leverage only pixel-wise masks while the remaining three methods have access to weak annotations (image-level labels).}
    \label{fig:qualitative}
\end{figure*}
}

\noindent\textbf{Bounding box supervision.}
In the lower half of Table~\ref{tab:mask_and_boundingbox}, we exchange pixel-wise masks with coarsely annotated bounding boxes.
As expected, the upper bound of full access at $39.39\%$ is limited by the nature of weak annotation.
Surprisingly, our \emph{Mean-Taught Deep Supervision} model having access to just $6$ \emph{pixel-wise masks} and additional global label (in the top half of Table~\ref{tab:mask_and_boundingbox}) already outperforms all models with full bounding box access.
Further, most of the best performing models build on  \emph{Multi-label Deep Supervision}.

\noindent\textbf{Results for other vendors.}
Due to the immense resources required to carry out the full evaluation as in Table~\ref{tab:mask_and_boundingbox}, i.e. training a total amount of $1100$ models, we show that our approaches generalize to other vendors (Cirrus and Topcon) on a smaller subset of experiments in Table~\ref{tab:vendors}.

\subsection{Qualitative results}
\noindent In~\figref{fig:qualitative}, we show segmentation results as more annotations are added for the \emph{Baseline} and \emph{Multi-label Deep Supervision} models as well as some semi-weakly supervised results.
Comparing the first two rows, we observe the advantageous effect of our deep supervision technique, especially in the low-data setting.
Similarly, when comparing our deep supervision in \emph{Mean-Taught} models (bottom row) to the rest, we directly see in column two and three that our models nicely segment the fluid regions, even with extremely few samples while other methods fail severely.

\section{Conclusion}
\noindent 
We introduced a new segmentation approach that allows for learning from different annotation types (pixel-wise labels, bounding boxes, global labels, unlabeled images). We showed that a key ingredient is the deep supervision principle, which we adapted to efficiently introduce supervision signals even in intermediate layers with a small spatial resolution. Furthermore, we introduced a student-teacher model to stabilize training and increase generalization.
Our experiments for a challenging medical application demonstrated the flexibility of our approach and the power of the simple concept of deep supervision for semi-weakly learning.

{\small
\bibliographystyle{ieee_fullname}
\bibliography{egbib}
}

\end{document}